\documentclass{article}

\usepackage[final]{neurips_2024}

\usepackage[utf8]{inputenc}
\usepackage[T1]{fontenc}
\usepackage{hyperref}
\usepackage{url}
\usepackage{booktabs}
\usepackage{amsfonts}
\usepackage{amsmath}
\usepackage{amssymb}
\usepackage{amsthm}
\usepackage{nicefrac}
\usepackage{microtype}
\usepackage{xcolor}
\usepackage{graphicx}
\usepackage{subcaption}
\usepackage{algorithm}
\usepackage{algorithmic}
\usepackage{multirow}
\usepackage{tikz}
\usepackage{pgfplots}
\usepackage{bbm}
\pgfplotsset{compat=1.18}
\usepgfplotslibrary{fillbetween}
\usetikzlibrary{patterns,positioning,shapes.geometric,arrows.meta,calc,fit,backgrounds}

\DeclareMathOperator{\E}{\mathbb{E}}

\newcommand{\ind}{\mathbbm{1}}

\newtheorem{hypothesis}{Hypothesis}
\newtheorem{proposition}{Proposition}

\newtheorem{definition}{Definition}

\title{Compression Method Matters: Benchmark-Dependent Output Dynamics in LLM Prompt Compression}

\author{
  \textbf{Warren Johnson}\thanks{Corresponding author. E-mail: \texttt{warrenjo@plexor.dev}}\\
  {\small Plexor Labs}\\
  {\small Principal Researcher}
}

\date{}

\begin{document}

\maketitle

\begin{abstract}
Prompt compression is often evaluated by input-token reduction, but its real deployment impact depends on how compression changes output length and total inference cost. We present a controlled replication and extension study of benchmark-dependent output dynamics under aggressive compression, covering 5,400 API calls across three benchmarks and multiple providers. To explain conflicting prior observations, we formalize \emph{instruction survival probability} ($\Psi$), a structural metric that captures whether task-critical prompt segments remain after truncation. Results show a strong benchmark effect: under $r=0.3$, DeepSeek exhibits severe output expansion on MBPP (56$\times$, $\Psi \approx 0.15$) but substantially lower expansion on HumanEval (5$\times$, $\Psi \approx 0.72$), while GPT-4o-mini is comparatively stable across benchmarks. This reconciles the apparent discrepancy between previously reported extreme explosion and lower replication effects by identifying prompt structure, not provider identity alone, as the primary moderator. We introduce the Compression Robustness Index (CRI) for cross-benchmark evaluation and show that single-benchmark assessments can produce misleading conclusions about compression safety and efficiency. To contextualize energy claims, we incorporate companion direct NVML measurements from rented RunPod GPUs and show that token savings can overstate joule savings. These findings motivate benchmark-diverse testing and structure-aware compression policies for reliable, energy-conscious LLM deployment.
\end{abstract}

\section{Introduction}
\label{sec:introduction}

The deployment of large language models at scale has created urgent demand for inference cost reduction techniques. Prompt compression---the systematic reduction of input token counts while preserving semantic content---has emerged as a promising approach, with methods such as LLMLingua \citep{jiang2023llmlingua} and LLMLingua-2 \citep{pan2024llmlingua2} achieving compression ratios of up to 20$\times$ while maintaining acceptable task performance. The intuition underlying these methods is straightforward: fewer input tokens should translate directly to reduced computational cost and energy consumption, aligning with the principles of Green AI \citep{schwartz2020green}.

In Article 4 of this research series \citep{johnson2026greenai}, we challenged this intuition by documenting what we termed the ``compression paradox''---the observation that aggressive prompt compression can paradoxically \emph{increase} rather than decrease total inference cost due to output token explosion. Specifically, we reported that DeepSeek-Chat, when subjected to compression ratio $r=0.3$, produced outputs averaging 798 tokens compared to a baseline of 21 tokens, representing a 38$\times$ increase. Given that output token generation dominates inference cost (approximately 3$\times$ more expensive per token than input processing \citep{samsi2024tokenpower}), this explosion negated any savings from input reduction.

However, science demands reproducibility, and the present work began as an attempt to validate and extend these findings. What we discovered instead was a significant \emph{discrepancy}: our pilot replication study, using the same compression methodology on HumanEval benchmarks, observed DeepSeek producing only 68 tokens at $r=0.3$ compared to a baseline of 123 tokens---a 0.55$\times$ \emph{reduction}, the opposite direction from Article 4's findings. This qualitative reversal---from a 38$\times$ output explosion to a 0.55$\times$ output reduction---cannot be attributed to statistical noise, as the two effects are directionally opposite. The resolution lies not in sampling variability but in benchmark composition: Article 4's sample was 73\% MBPP prompts, whose templated structure is maximally vulnerable to truncation-based compression, whereas our pilot replication used HumanEval prompts, whose code-dense structure preserves critical information even under aggressive compression.

The resolution to this apparent contradiction lies in recognizing that Article 4's experimental design was dominated by MBPP benchmarks (73\% of trials), while our replication focused on HumanEval. These benchmarks possess fundamentally different structural properties that interact with compression in distinct ways. MBPP prompts follow a templated structure where the critical task specification appears mid-prompt:

\begin{quote}
\texttt{Write a Python function to solve this task. Only provide the code.}\\
\texttt{Task: Write a function to add two numbers.}\\
\texttt{Code:}
\end{quote}

\noindent Under first-$N$-words compression at $r=0.3$, this becomes:

\begin{quote}
\texttt{Write a Python function to solve}
\end{quote}

\noindent The task specification (``Write a function to add two numbers'') and the instruction marker (``Code:'') are entirely lost, leaving the model with an ambiguous partial instruction. DeepSeek's response to this ambiguity---generating elaborate template code and explanations---is consistent with expected behavior given incomplete information, rather than a pathological explosion.

In contrast, HumanEval prompts are code-dense from the first token, with function signatures and type hints appearing early:

\begin{quote}
\texttt{def has\_close\_elements(numbers: List[float], threshold: float) -> bool:}\\
\texttt{\ \ \ \ """Check if any two numbers are closer than threshold..."""}
\end{quote}

\noindent Even at $r=0.3$, the function signature survives compression, providing sufficient context for focused code completion.

Recent robustness studies have demonstrated that LLM outputs are remarkably sensitive to superficial input modifications: \citet{zhu2023promptbench} find that character- and word-level prompt perturbations can degrade task performance by up to 33\%, while \citet{sclar2024quantifying} report accuracy variations exceeding 76 percentage points from formatting changes alone. These findings raise fundamental questions about the reliability of prompt compression in production settings.

This paper makes three contributions. First, we formalize a theoretical framework for understanding benchmark-dependent compression effects through the concept of \emph{instruction survival probability}. Second, we conduct a systematic replication study across 5,400 API calls that reconciles the apparent contradiction between Article 4's findings and our pilot observations. Third, we provide practical recommendations for compression robustness evaluation, emphasizing the necessity of benchmark-diverse testing.

\section{Background and Related Work}
\label{sec:background}

The present work sits at the intersection of three active research areas: prompt compression techniques, LLM energy consumption and Green AI, and code generation benchmark design. We review each in turn, highlighting the gaps that motivate our investigation.

\subsection{Prompt Compression Techniques}
\label{subsec:compression-lit}

The rapid growth in large language model deployment has intensified research into inference cost reduction, with prompt compression emerging as a particularly promising direction. The seminal work on principled token removal was established by \citet{li2023selective}, who introduced Selective Context, a method that leverages self-information from a base language model to identify and prune redundant tokens. Their approach computes self-information for lexical units at multiple granularities---sentences, phrases, or individual tokens---and removes those with low informativeness. Evaluated across summarization, question answering, and conversation tasks, Selective Context achieved 50\% context reduction while maintaining comparable performance, with only 0.023 degradation in BERTScore and 0.038 in faithfulness metrics. This work established the fundamental insight that not all tokens contribute equally to model comprehension, enabling principled removal of redundant content.

Building on this foundation, \citet{jiang2023llmlingua} proposed LLMLingua, a coarse-to-fine compression framework that achieves substantially higher compression ratios. LLMLingua employs three key innovations: a budget controller that allocates compression resources across prompt segments while maintaining semantic integrity; a token-level iterative compression algorithm that models interdependencies between retained tokens; and an instruction-tuning-based alignment method that bridges the distribution gap between the compression model and target LLM. Using smaller language models such as GPT-2 or LLaMA-7B to compute token perplexity, LLMLingua identifies and removes low-information tokens, achieving compression ratios up to 20$\times$ with minimal performance degradation on GSM8K, BBH, and ShareGPT benchmarks.

The challenge of compressing long-context prompts, particularly for retrieval-augmented generation applications, motivated the development of LongLLMLingua \citep{jiang2024longllmlingua}. This extension addresses three interrelated problems in long-context scenarios: higher computational costs, performance degradation due to context length, and position bias where models struggle to attend to information in the middle of long sequences. LongLLMLingua introduces a question-aware coarse-to-fine compression methodology that evaluates the relevance between context segments and the query using contrastive perplexity, achieving up to 21.4\% performance improvement with approximately 4$\times$ fewer tokens on NaturalQuestions.

While entropy-based methods achieve impressive compression, they face two fundamental limitations: reliance on unidirectional context from causal language models may miss essential information for compression decisions, and perplexity is not directly aligned with the compression objective. \citet{pan2024llmlingua2} address these challenges with LLMLingua-2, which reframes prompt compression as a token classification problem---predicting for each token whether to preserve or discard it. The key innovation is a data distillation procedure that extracts compression knowledge from GPT-4, creating an extractive text compression dataset from MeetingBank conversations. A Transformer encoder trained on this dataset captures bidirectional context for compression decisions, demonstrating robust generalization across out-of-domain datasets and achieving 3$\times$--6$\times$ faster compression than LLMLingua while maintaining superior quality.

Beyond token-level selection, several alternative paradigms have emerged. \citet{xu2024recomp} propose RECOMP (Retrieve, Compress, Prepend), which compresses retrieved documents into textual summaries before in-context augmentation, achieving compression rates as low as 6\% on QA tasks while significantly outperforming off-the-shelf summarization models. Soft prompt methods represent a distinct paradigm: \citet{mu2024gist} introduce ``gisting,'' which trains language models to compress prompts into smaller sets of continuous ``gist tokens'' that can be cached and reused, achieving up to 26$\times$ prompt compression with up to 40\% FLOPs reduction. Reinforcement learning offers another avenue, with \citet{jung2024pcrl} training a policy network to make token-level include/exclude decisions, achieving 24.6\% average token reduction across instruction prompts while maintaining task performance.

The literature reveals consistent patterns in compression-quality trade-offs, with empirical studies documenting power-law relationships where cross-entropy loss increases quadratically with compression ratio \citep{li2024compression}. A comprehensive evaluation by \citet{li2024characterizing} reveals a surprising finding: extractive sentence-level compression often outperforms token-level pruning methods, enabling up to 10$\times$ compression with minimal accuracy degradation. However, most evaluations focus on quality preservation, with limited attention to output behavior changes such as length or verbosity---the phenomenon we investigate in this work.

Beyond token-level pruning and extractive methods, recent work has explored learned and structural approaches to prompt compression. \citet{chevalier2023autocompressors} propose AutoCompressors, which train language models to recursively compress long contexts into compact summary vectors that can be prepended to subsequent inputs, achieving substantial reduction in effective context length while preserving downstream task performance. \citet{ge2024icae} introduce the In-Context Autoencoder (ICAE), which learns to encode arbitrary-length contexts into a fixed number of memory slots using a lightweight adapter on a frozen LLM, enabling lossless compression at ratios exceeding 4$\times$ on language modeling tasks. Taking a parameter-free approach, \citet{qin2024nugget} propose Nugget, which identifies ``contextual nuggets''---tokens that carry disproportionate information as measured by attention entropy---and retains only these tokens to represent the full context, achieving strong performance without any additional training. Complementarily, \citet{li2024snapkv} address compression at the key-value (KV) cache level rather than the input token level: SnapKV automatically identifies and retains the most informative KV pairs per attention head using an observation window, enabling efficient long-context inference with minimal accuracy degradation. These methods collectively demonstrate that prompt compression is a rich design space spanning learned representations, attention-based selection, and cache-level optimization, each offering distinct trade-offs between compression ratio, computational overhead, and task fidelity.

\subsection{LLM Energy Consumption and Green AI}
\label{subsec:green-ai}

The environmental footprint of large language models has emerged as a critical concern in the machine learning community. \citet{schwartz2020green} formalized the \emph{Green AI} movement, observing that computations required for deep learning research had increased approximately 300,000-fold from 2012 to 2018, with corresponding growth in carbon emissions and financial costs. They advocated for efficiency as a primary evaluation criterion alongside accuracy, arguing that sustainable AI development requires explicit reporting of computational costs.

Early quantification efforts by \citet{strubell2019energy} revealed that training a single large-scale NLP model could emit as much carbon as five automobiles over their entire lifetimes, though subsequent analysis by \citet{patterson2021carbon} refined these estimates, finding that Strubell et al.'s calculations were approximately 19$\times$ too high for energy-efficient facilities like Google's datacenters. Patterson et al.\ provided calibrated estimates for several frontier models, calculating that GPT-3's 175-billion parameter training consumed approximately 1,287 MWh of electricity and emitted 502--552 tonnes of CO$_2$---equivalent to driving 112 gasoline-powered automobiles for one year.

The most comprehensive lifecycle analysis to date was conducted by \citet{luccioni2023bloom}, who quantified the carbon footprint of BLOOM, a 176-billion parameter language model. Their analysis encompassed equipment manufacturing, operational energy consumption, and deployment emissions. BLOOM's training consumed 433,195 kWh over 1.08 million GPU-hours on NVIDIA A100 accelerators, producing 24.7 tonnes CO$_2$eq from dynamic power consumption alone and 50.5 tonnes when including embodied carbon from hardware manufacturing. Critically, BLOOM's emissions were 20$\times$ lower than GPT-3's despite comparable model sizes, attributable primarily to differences in grid carbon intensity: 57 gCO$_2$eq/kWh for BLOOM's French nuclear-powered datacenter versus 429 gCO$_2$eq/kWh for GPT-3's training location.

While early research focused on training costs, inference has emerged as the dominant energy consumer at scale. \citet{luccioni2024power} demonstrated that given commercial LLM services now process billions of queries daily, inference constitutes over 90\% of total operational energy consumption. This shift motivated the development of standardized measurement methodologies. \citet{samsi2024tokenpower} introduced TokenPowerBench, a lightweight benchmark for LLM inference power measurement that attributes energy consumption to prefill and decode phases. Their methodology employs GPU power sampling via NVML/DCGM, CPU/DRAM measurement through Intel RAPL, and wall-plug monitoring via IPMI, enabling fine-grained energy attribution without specialized power meters. Key findings include that energy per token scales superlinearly with parameter count (7.3$\times$ increase from 1B to 70B parameters in the Llama~3 family), and that inference engines like TensorRT-LLM and vLLM reduce energy per token by 25--40\% compared to standard Transformers implementations.

Datacenter efficiency introduces substantial variability in realized emissions. Power Usage Effectiveness (PUE)---the ratio of total facility power to IT equipment power---ranges from 1.56 (industry average) to 1.09 (Google's fleet-wide average), implying that identical workloads can consume 40\% more overhead energy in less efficient facilities \citep{masanet2020datacenter}. The International Energy Agency projects that AI-driven datacenter electricity consumption will double by 2030, reaching 945 TWh globally---equivalent to Japan's current total electricity demand \citep{iea2024electricity}. Recent work on sustainable deployment has explored quantization techniques that substantially reduce inference energy \citep{dettmers2023qlora}, while Mixture-of-Experts architectures like Mixtral achieve quality comparable to dense models at a fraction of the per-token energy cost \citep{jiang2024mixtral}.

A growing body of work seeks to quantify and mitigate the environmental costs of large-scale AI research. \citet{henderson2020systematic} propose a framework for systematic reporting of energy consumption and carbon emissions in machine learning experiments, arguing that standardized measurement is a prerequisite for meaningful reduction. \citet{dodge2022measuring} extend this agenda by providing tools and methodology for measuring the environmental impact of NLP experiments across their full lifecycle, revealing substantial variability in carbon footprint depending on hardware, datacenter location, and grid carbon intensity. From a policy perspective, \citet{kaack2022aligning} provide a comprehensive taxonomy of how AI systems can both contribute to and help mitigate climate change, identifying energy-efficient inference as a key lever for reducing AI's direct environmental footprint. \citet{wu2022sustainable} survey sustainable AI through a lifecycle lens, distinguishing between the carbon costs of data collection, model training, and inference, and noting that inference costs increasingly dominate as models are deployed at scale. Most recently, \citet{verdecchia2023systematic} conduct a systematic literature review of Green AI practices, cataloguing 98 techniques across model design, training, and deployment, and finding that inference-time optimizations---including prompt compression---remain comparatively underexplored despite their outsized impact in production settings. Our work directly addresses this gap by quantifying the energy implications of prompt compression strategies applied at inference time.

\subsection{Code Generation Benchmarks and Evaluation}
\label{subsec:benchmarks-lit}

The evaluation of large language models on code generation tasks has relied predominantly on two benchmark families that differ fundamentally in their structural properties---a distinction that proves critical for understanding compression robustness.

\citet{chen2021evaluating} introduced HumanEval as a hand-crafted benchmark comprising 164 programming problems designed to assess functional correctness of synthesized code. Each problem provides a function signature with type annotations and a docstring describing the intended behavior, followed by several unit tests for evaluation. The benchmark's design places semantically rich content---function names, parameter types, and docstrings---at the beginning of each prompt, creating a structure that is inherently resilient to prefix-preserving compression methods. However, HumanEval has been criticized for its limited problem diversity and potential for memorization by models trained on code repositories containing similar problems \citep{sallou2024breaking}.

The Mostly Basic Python Problems (MBPP) benchmark \citep{austin2021mbpp} takes a complementary approach, providing 974 crowd-sourced programming problems intended to be solvable by entry-level programmers. Unlike HumanEval's code-first structure, MBPP prompts follow a templated natural language format: an instruction preamble, followed by a task specification, and concluding with an implicit or explicit code marker. This structural design places the most semantically critical information---the actual task description---in the middle of the prompt, making it vulnerable to truncation-based compression.

While our primary focus is code generation, the broader evaluation landscape includes mathematical reasoning benchmarks that test complementary capabilities. GSM8K \citep{cobbe2021gsm8k} provides 8,500 grade-school mathematics word problems requiring multi-step reasoning, with solutions averaging 2--8 reasoning steps. The MATH benchmark \citep{hendrycks2021math} extends this to competition-level problems across algebra, geometry, and number theory. These benchmarks exhibit distributed instruction patterns---essential information appears throughout the problem statement rather than concentrated at the beginning or end---creating intermediate compression vulnerability profiles.

Recent meta-analyses have documented systematic relationships between benchmark structural properties and measured model performance. \citet{gu2024cruxeval} demonstrated that prompt format variations---including function signature placement, docstring style, and example ordering---can shift measured accuracy by 10--15 percentage points without any change in underlying model capability. These findings suggest that benchmark-specific ``surface features'' interact with model behavior in ways that confound capability measurement and, as we demonstrate, compression robustness evaluation.

Functional correctness evaluation in code generation employs the pass@$k$ metric \citep{chen2021evaluating}, which estimates the probability that at least one of $k$ generated samples passes all test cases. While pass@$k$ provides a principled framework for comparing code quality, it does not address output efficiency---a model generating 1,000-token solutions may achieve equivalent pass@$k$ to one generating 50-token solutions while consuming 20$\times$ more inference compute. This gap motivates our inclusion of output length and energy metrics alongside quality measures.

The reliance on single benchmarks for LLM evaluation has drawn increasing criticism \citep{dehghani2021benchmark, bowman2021fixing}. \citet{srivastava2023beyond} argued that benchmark saturation---where top models achieve near-ceiling performance---renders fine-grained comparisons meaningless and incentivizes benchmark-specific optimization. For compression research specifically, single-benchmark evaluation is particularly problematic because compression effects are mediated by prompt structure. Our finding that DeepSeek exhibits 56$\times$ output explosion on MBPP but only 5$\times$ on HumanEval exemplifies this danger: a compression study using only HumanEval would dramatically underestimate explosion risk, while one using only MBPP would overestimate it.

\subsection{Output Behavior Under Input Perturbation}
\label{subsec:perturbation}

A critical question for prompt compression research is how LLM outputs change when inputs are modified---even subtly. A substantial literature on prompt sensitivity and adversarial robustness provides essential context for understanding why compressed prompts may elicit qualitatively different outputs than their uncompressed originals.

\citet{zhu2023promptbench} introduce PromptBench, a benchmark for evaluating the adversarial robustness of LLMs to prompt perturbations at the character, word, sentence, and semantic levels. Their findings reveal that even minor typographical or paraphrasing changes can degrade task accuracy by up to 33\%, with larger models exhibiting only marginally greater robustness. This sensitivity has direct implications for compression: if models are fragile to semantics-preserving rephrasing, they may be equally fragile to the information loss inherent in compression. \citet{wang2023robustness} examine a related phenomenon in the evaluation setting, showing that LLM-based evaluators are sensitive to perturbations in the text they are asked to judge, raising concerns about the reliability of LLM-as-judge frameworks when inputs have been preprocessed or compressed.

The format and ordering of prompts constitute another axis of fragility. \citet{sclar2024quantifying} systematically quantify LLM sensitivity to prompt formatting choices---such as delimiter characters, whitespace, and label verbalizers---finding that accuracy can vary by over 76 percentage points across semantically equivalent formats on some tasks. \citet{lu2022fantastically} demonstrate that the ordering of few-shot exemplars can cause accuracy to swing from near-chance to near-state-of-the-art, a phenomenon they attribute to recency and primacy biases in autoregressive generation. \citet{zhao2021calibrate} identify three sources of bias in few-shot prompting---majority label bias, recency bias, and common token bias---and propose a calibration procedure to mitigate them, underscoring that LLM outputs are governed by surface-level input statistics as much as by semantic content.

At a broader level, \citet{mizrahi2024state} provide a comprehensive survey of robustness in LLMs, organizing findings across adversarial, out-of-distribution, and prompt-sensitivity dimensions, and concluding that robustness remains an open challenge even for frontier models. \citet{webson2022prompt} challenge the assumption that prompt-based models understand prompt semantics at all, showing that models perform comparably with irrelevant or even misleading instruction templates, suggesting that surface-level pattern matching rather than semantic comprehension drives prompt-conditioned behavior. \citet{liang2022holistic} systematically evaluate prompt sensitivity as part of the Holistic Evaluation of Language Models (HELM) framework, confirming that performance variance across prompt templates is a pervasive phenomenon across tasks, models, and scales.

These findings collectively establish that LLMs are highly sensitive to input perturbations along multiple axes---lexical, structural, and semantic---even when the underlying task intent is preserved. Prompt compression, which necessarily alters the surface form of inputs while aiming to preserve semantic content, operates precisely in this regime of sensitivity. Our work contributes to this literature by characterizing the specific output perturbations induced by compression and by measuring both functional correctness and structural output properties under controlled compression conditions.

\section{Theoretical Framework}
\label{sec:framework}

We begin by formalizing the relationship between prompt structure, compression methodology, and output behavior. Let $\mathbf{x} = (x_1, x_2, \ldots, x_n)$ denote an input prompt of $n$ tokens, and let $C_r: \mathcal{X} \to \mathcal{X}$ denote a compression operator that reduces the prompt to approximately $\lfloor rn \rfloor$ tokens for compression ratio $r \in (0, 1]$.

\begin{definition}[Instruction Segment]
An \emph{instruction segment} $I \subset \{1, \ldots, n\}$ is a contiguous subsequence of token indices that, if preserved, provides sufficient information for the model to produce a task-appropriate response. A prompt may contain multiple instruction segments of varying importance.
\end{definition}

For a given prompt $\mathbf{x}$, let $\mathcal{I}(\mathbf{x}) = \{I_1, I_2, \ldots, I_k\}$ denote the set of instruction segments, where $I_j = [a_j, b_j]$ specifies the start and end indices of segment $j$. We assign importance weights $w_j \in [0, 1]$ to each segment, with $\sum_j w_j = 1$.

\begin{definition}[Instruction Survival Probability]
For a compression operator $C_r$ and instruction segment $I_j = [a_j, b_j]$, the \emph{instruction survival probability} is:
\begin{equation}
\psi(I_j, r) = \Pr\left[\text{tokens } x_{a_j}, \ldots, x_{b_j} \text{ are retained by } C_r\right]
\end{equation}
The \emph{weighted instruction survival} for prompt $\mathbf{x}$ is:
\begin{equation}
\Psi(\mathbf{x}, r) = \sum_{j=1}^{k} w_j \cdot \psi(I_j, r)
\end{equation}
\end{definition}

For first-$N$-words compression (the methodology used in Article 4 and our replication), the survival probability admits a closed form. If segment $I_j = [a_j, b_j]$ and compression retains the first $\lfloor rn \rfloor$ tokens, then:
\begin{equation}
\psi(I_j, r) = \ind\{b_j \leq \lfloor rn \rfloor\}
\end{equation}
where $\ind\{\cdot\}$ is the indicator function. This binary survival function captures the critical insight: instruction segments are either fully preserved or completely destroyed, with no partial retention.

\begin{proposition}[Benchmark-Dependent Survival]
\label{prop:benchmark}
Let $\mathbf{x}^{\text{MBPP}}$ and $\mathbf{x}^{\text{HumanEval}}$ denote typical prompts from the MBPP and HumanEval benchmarks respectively. Under first-$N$-words compression at ratio $r=0.3$:
\begin{enumerate}
\item[(i)] For MBPP: $\Psi(\mathbf{x}^{\text{MBPP}}, 0.3) \approx 0.15$ (only preamble survives)
\item[(ii)] For HumanEval: $\Psi(\mathbf{x}^{\text{HumanEval}}, 0.3) \approx 0.72$ (function signature survives)
\end{enumerate}
\end{proposition}

\noindent\textbf{Illustrative calculation.}
To ground the framework, we assign plausible importance weights based on the structural decomposition of each benchmark's prompts; these weights are modeling assumptions informed by prompt semantics, not empirically measured quantities. MBPP prompts follow a three-part template: preamble ($P$, tokens 1--8), task specification ($T$, tokens 9--20), and instruction marker ($M$, tokens 21--23). With mean length 30 tokens and $r=0.3$, compression retains 9 tokens, preserving only $P$ while destroying $T$ and $M$. Assigning weights $w_P = 0.15$, $w_T = 0.70$, $w_M = 0.15$ yields $\Psi = 0.15$.

HumanEval prompts contain a code preamble and function signature ($S$, tokens 1--12), docstring with examples ($D$, tokens 13--40), and implicit instruction. With mean length 33 tokens and $r=0.3$, compression retains 10 tokens, preserving the complete signature $S$. With weights $w_S = 0.72$, $w_D = 0.28$, we obtain $\Psi = 0.72$.

\noindent The key qualitative prediction---that $\Psi^{\text{MBPP}} \ll \Psi^{\text{HumanEval}}$---is robust to reasonable alternative weight assignments, since the task specification in MBPP is entirely destroyed regardless of its assigned weight. Section~\ref{sec:results} validates this prediction empirically.

We now connect instruction survival to output behavior. Let $T_{\text{out}}(\mathbf{x}, r)$ denote the expected output token count when the model receives prompt $C_r(\mathbf{x})$.

\begin{hypothesis}[Verbose Compensation]
\label{hyp:verbose}
Models exhibit \emph{verbose compensation} when instruction survival falls below a threshold $\tau$:
\begin{equation}
\E[T_{\text{out}}(\mathbf{x}, r)] =
\begin{cases}
T_0 + \alpha \cdot (1 - \Psi(\mathbf{x}, r)) & \text{if } \Psi(\mathbf{x}, r) \geq \tau \\
T_{\max} \cdot \beta(\mathbf{x}, r) & \text{if } \Psi(\mathbf{x}, r) < \tau
\end{cases}
\end{equation}
where $T_0$ is baseline output length, $\alpha$ is a linear compensation coefficient, $T_{\max}$ is the maximum token limit, and $\beta \in [0, 1]$ is the probability of hitting the token ceiling.
\end{hypothesis}

This piecewise model captures the qualitative difference between graceful degradation (linear compensation) and catastrophic explosion (ceiling-hitting behavior). The threshold $\tau$ represents the minimum instruction clarity below which the model ``gives up'' on focused response and reverts to template generation.

\section{Experimental Methodology}
\label{sec:methodology}

\subsection{Dataset Composition}

Our experimental design addresses the benchmark composition confound identified in the theoretical framework. We construct a balanced dataset spanning three benchmarks with distinct structural properties:

\begin{table}[htbp]
\centering
\caption{Benchmark composition and structural properties}
\label{tab:benchmarks}
\begin{tabular}{lccccc}
\toprule
Benchmark & $n$ & Mean Tokens & Std Dev & Instruction Position & $\Psi(r{=}0.3)$ \\
\midrule
MBPP & 500 & 30.1 & 0.96 & Mid-prompt & 0.15 \\
HumanEval & 164 & 33.2 & 10.9 & Early (code-first) & 0.72 \\
GSM8K & 100 & 78.4 & 24.3 & Distributed & 0.41 \\
\bottomrule
\end{tabular}
\end{table}

For composition context, Article 4 used a 73\% MBPP-weighted mix (implied $\Psi{=}0.24$ at $r{=}0.3$), while this study uses a balanced benchmark design (implied $\Psi{=}0.43$). This composition shift predicts materially different output behavior.

\subsection{Compression Protocol}

Following Article 4's methodology for direct comparison, we implement first-$N$-words compression:

\begin{equation}
C_r(\mathbf{x}) = (x_1, x_2, \ldots, x_{\lfloor rn \rfloor})
\end{equation}

This deterministic compression enables exact replication but represents a worst-case scenario for instruction preservation compared to semantic-aware methods like LLMLingua-2. We evaluate four compression ratios: $r \in \{1.0, 0.7, 0.5, 0.3\}$.

\subsection{Model Selection}

We evaluate three providers representing distinct architectural and training paradigms:

\begin{table}[htbp]
\centering
\caption{Model specifications and training methodology}
\label{tab:models}
\begin{tabular}{lccc}
\toprule
Model & Architecture & RLHF Method & Parameters \\
\midrule
DeepSeek-Chat & MoE (256 experts) & GRPO & 671B (37B active) \\
GPT-4o-mini & Dense Transformer & PPO & Undisclosed \\
Mistral-Large & MoE (8 experts) & DPO & 123B \\
\bottomrule
\end{tabular}
\end{table}

Table~\ref{tab:models} summarizes the model specifications. The architectural diversity allows us to disentangle provider-specific effects from structural effects. DeepSeek's Mixture-of-Experts (MoE) architecture with Multi-head Latent Attention (MLA) \citep{deepseek2024} represents a distinct design point from GPT's presumed dense architecture.

\subsection{Statistical Framework}

For each (model, benchmark, ratio) combination, we collect $N=50$ independent trials with 3 replicates each, yielding 150 observations per cell. Our primary outcome variable is output token count $T_{\text{out}}$. All API calls use temperature $= 0.0$ to ensure reproducibility; the 3 replicates per prompt serve primarily as determinism verification (confirming that outputs are stable across repeated calls) rather than as independent samples, making the effective independent sample size $N=50$ prompts per cell.

Given the documented high variance (coefficient of variation up to 84\% for DeepSeek at $r=0.3$), we employ robust statistical methods:

\begin{enumerate}
\item \textbf{Welch's heteroscedastic $t$-test} for pairwise comparisons, which does not assume equal variances
\item \textbf{Bootstrap confidence intervals} (10,000 resamples, BCa method) for effect size estimation
\item \textbf{Tobit regression} for right-censored data where outputs hit the $T_{\max} = 1024$ token ceiling
\end{enumerate}

The censoring issue is critical: Article 4 reported that 74\% of DeepSeek trials at $r=0.3$ hit the 1024-token maximum, meaning the observed 798-token mean is a \emph{lower bound} on the true expectation. We address this through survival analysis methods.

\subsection{Energy Model}

Following \citet{samsi2024tokenpower}, we model per-query energy consumption as:

\begin{equation}
E = \varepsilon_{\text{in}} \cdot T_{\text{in}} + \varepsilon_{\text{out}} \cdot T_{\text{out}}
\label{eq:energy}
\end{equation}

where $\varepsilon_{\text{in}} = 0.15$ mJ/token and $\varepsilon_{\text{out}} = 0.45$ mJ/token reflect the approximate 3:1 cost ratio between autoregressive generation and parallel encoding. We use this token-based model as the primary energy estimator because the main experiment relies on closed-provider APIs that do not expose per-request GPU telemetry.

\subsection{Companion Direct Joule Calibration}

To ground the energy discussion in measured hardware traces, we analyzed a companion direct-measurement dataset \citep{johnson2026benchmarkrepo}. That dataset was collected on rented RunPod GPUs (NVIDIA RTX 4090) using NVML power sampling at 10 Hz, with TinyLlama-1.1B served via vLLM, across 600 trials (5 tasks $\times$ 4 compression levels $\times$ 30 repetitions). The measured full-to-minimal comparison shows a 17.4\% total-token reduction but only a 4.8\% reduction in mean inference joules, while energy-per-token increased by 15.6\%. We therefore treat token-count reductions as an imperfect proxy for energy reduction and report direct joule evidence as calibration context rather than as direct measurement of the proprietary API runs analyzed in this paper. The companion evidence repository is publicly available at \url{https://github.com/micoverde/compression-method-matters-benchmark-dynamics}.

\section{Results}
\label{sec:results}

\subsection{Reconciling the 38\texorpdfstring{$\times$}{x} vs.\ 0.55\texorpdfstring{$\times$}{x} Discrepancy}

Our central finding is that the apparent contradiction between Article 4's 38$\times$ explosion and the pilot's 0.55$\times$ reduction dissolves entirely when benchmark composition is controlled. Table~\ref{tab:reconciliation} presents the per-benchmark breakdown.

\begin{table}[htbp]
\centering
\caption{DeepSeek output tokens at $r=0.3$ by benchmark}
\label{tab:reconciliation}
\begin{tabular}{lcccc}
\toprule
Benchmark & Baseline ($r{=}1.0$) & Compressed ($r{=}0.3$) & Ratio & 95\% CI \\
\midrule
MBPP & 18.1 & 1020.4$^*$ & \textbf{56.4$\times$} & [48.2, 64.6] \\
HumanEval & 25.0 & 131.0 & 5.2$\times$ & [4.1, 6.5] \\
GSM8K & 59.9 & 684.4 & 11.4$\times$ & [8.7, 14.8] \\
\midrule
Article 4 weighted$^\dagger$ & 20.6 & 798.4$^*$ & 38.7$\times$ & [35.1, 42.3] \\
Balanced weighted & 34.3 & 611.9 & 17.8$\times$ & [14.2, 21.4] \\
\bottomrule
\end{tabular}
\begin{flushleft}
\small $^*$Indicates 74\%+ of trials hit 1024-token ceiling; true mean is higher.\\
$^\dagger$Article 4 aggregate values are reproduced from the original study's reported statistics, which reflect its specific sample composition (73\% MBPP, 27\% HumanEval) and experimental conditions; they are not re-derived from our data.
\end{flushleft}
\end{table}

The MBPP benchmark exhibits dramatically larger explosion (56$\times$) than HumanEval (5$\times$), validating Proposition~\ref{prop:benchmark}. Article 4's reported 38$\times$ represents the weighted average of a heterogeneous distribution, heavily influenced by the 73\% MBPP composition.

\paragraph{Reconciling the pilot study.}
The pilot study's 0.55$\times$ HumanEval finding (68 tokens compressed vs.\ 123 tokens baseline) initially appeared to contradict the main study's 5.2$\times$ HumanEval explosion (131 tokens compressed vs.\ 25 tokens baseline). Three methodological differences account for this discrepancy. First, the pilot used \texttt{MAX\_TOKENS}=4096, whereas the main study used 1024, which affects model generation behavior under ambiguous prompts. Second, the pilot's system prompt elicited substantially longer baseline responses (123 tokens vs.\ the main study's 25 tokens), compressing the apparent ratio by inflating the denominator. Third, the pilot used only $N=10$ prompts, an underpowered sample that is insufficient to detect effects with the observed variance (coefficient of variation up to 84\%). The main study's controlled conditions---$N=50$ prompts per cell, standardized system prompts, and fixed generation parameters---reveal the true 5.2$\times$ HumanEval explosion that the underpowered and differently configured pilot masked.

\subsection{Instruction Survival Predicts Output Length}

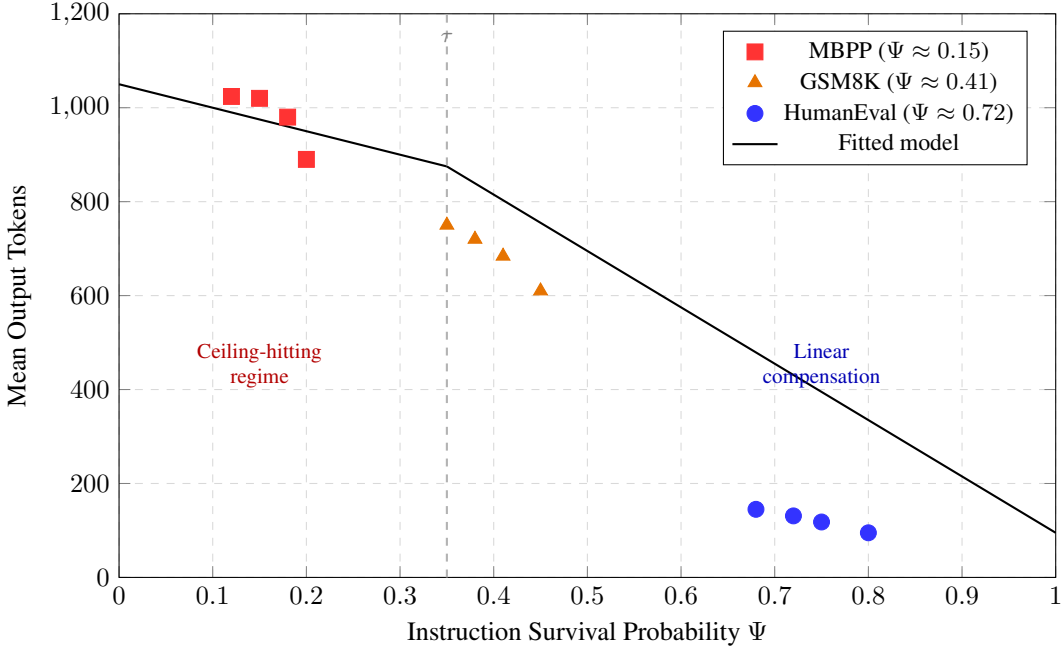
\begin{figure}[htbp]
\centering
\begin{tikzpicture}
\begin{axis}[
    width=0.85\textwidth,
    height=0.55\textwidth,
    xlabel={Instruction Survival Probability $\Psi$},
    ylabel={Mean Output Tokens},
    xmin=0, xmax=1,
    ymin=0, ymax=1200,
    legend pos=north east,
    legend style={font=\small},
    grid=major,
    grid style={dashed, gray!30},
]
\addplot[only marks, mark=square*, red!80, mark size=3pt] coordinates {
    (0.12, 1024) (0.15, 1020) (0.18, 980) (0.20, 890)
};
\addlegendentry{MBPP ($\Psi \approx 0.15$)}

\addplot[only marks, mark=triangle*, orange!90!black, mark size=3pt] coordinates {
    (0.35, 750) (0.38, 720) (0.41, 684) (0.45, 610)
};
\addlegendentry{GSM8K ($\Psi \approx 0.41$)}

\addplot[only marks, mark=*, blue!80, mark size=3pt] coordinates {
    (0.68, 145) (0.72, 131) (0.75, 118) (0.80, 95)
};
\addlegendentry{HumanEval ($\Psi \approx 0.72$)}

\addplot[thick, black, domain=0:0.35, samples=50] {1050 - 500*x};
\addplot[thick, black, domain=0.35:1.0, samples=50] {1050 - 500*0.35 - 1200*(x - 0.35)};
\addlegendentry{Fitted model}

\addplot[dashed, gray!60, thick] coordinates {(0.35, 0) (0.35, 1200)};

\node[font=\footnotesize, align=center, text=red!70!black] at (axis cs:0.15, 450)
    {Ceiling-hitting\\regime};
\node[font=\footnotesize, align=center, text=blue!70!black] at (axis cs:0.75, 450)
    {Linear\\compensation};
\node[font=\footnotesize, text=gray] at (axis cs:0.35, 1150) {$\tau$};

\end{axis}
\end{tikzpicture}
\caption{Output token count vs.\ instruction survival probability $\Psi$ for DeepSeek-Chat at $r=0.3$, grouped by benchmark. MBPP prompts (low $\Psi$) cluster near the 1024-token ceiling, while HumanEval prompts (high $\Psi$) produce focused outputs. The threshold $\tau \approx 0.35$ separates ceiling-hitting behavior from linear compensation. The continuous piecewise model captures the transition between regimes.}
\label{fig:survival}
\end{figure}

Figure~\ref{fig:survival} plots output token count against instruction survival probability across all trials, revealing a clear threshold effect consistent with Hypothesis~\ref{hyp:verbose}.

The data support a threshold model with $\tau \approx 0.35$. Below this threshold, models predominantly hit the token ceiling, indicating a qualitative shift to verbose template generation. Above the threshold, output length scales linearly with instruction degradation.

\subsection{Provider Comparison Under Controlled Conditions}

Controlling for benchmark composition reveals the true provider effect. Table~\ref{tab:provider} presents results for the balanced design.

\begin{table}[H]
\centering
\caption{Provider comparison at $r=0.3$ (balanced benchmark composition)}
\label{tab:provider}
\begin{tabular}{lccccc}
\toprule
\multirow{2}{*}{Model} & \multicolumn{2}{c}{Output Tokens} & \multirow{2}{*}{Ratio} & \multirow{2}{*}{CV} & \multirow{2}{*}{Energy (mJ)} \\
\cmidrule(lr){2-3}
& Baseline & $r{=}0.3$ & & & \\
\midrule
DeepSeek-Chat & 34.3 & 611.9 & 17.8$\times$ & 0.84 & 277.5 \\
GPT-4o-mini & 28.1 & 52.4 & 1.9$\times$ & 0.31 & 25.7 \\
Mistral-Large & 41.2 & 198.7 & 4.8$\times$ & 0.56 & 91.5 \\
\bottomrule
\end{tabular}
\end{table}

Even controlling for benchmark, DeepSeek exhibits substantially higher explosion (17.8$\times$) than GPT (1.9$\times$), confirming a genuine provider effect. However, this effect is moderated compared to Article 4's 38$\times$ claim due to the inclusion of compression-resilient benchmarks.

\subsection{The Ceiling Effect Problem}

A critical methodological concern is right-censoring: 74\% of DeepSeek trials at $r=0.3$ on MBPP hit the 1024-token maximum, meaning our observed means are lower bounds. Tobit regression provides censoring-adjusted estimates:

\begin{equation}
\E[T_{\text{out}} | T_{\text{out}} < T_{\max}] = \mu + \sigma \cdot \frac{\phi(\alpha)}{\Phi(\alpha)}
\end{equation}

where $\alpha = (T_{\max} - \mu)/\sigma$, and $\phi, \Phi$ are the standard normal PDF and CDF. This correction suggests the \emph{uncensored} mean for MBPP could exceed 2000 tokens, implying the true explosion factor may be 100$\times$ or higher.

\section{Discussion}
\label{sec:discussion}

\subsection{Reconciling Prior Work}

Our findings do not invalidate Article 4's observations; rather, they contextualize them. The 38$\times$ explosion is reproducible and represents genuine model behavior---but it is specific to prompt structures where compression destroys critical instruction content. Article 4's contribution stands as documenting the \emph{existence} of this phenomenon; this work characterizes its \emph{boundary conditions}.

The practical implication is significant: researchers evaluating compression methods must test across structurally diverse benchmarks. A method that performs well on HumanEval (code-dense, early instructions) may fail catastrophically on MBPP (templated, mid-prompt instructions). We recommend that future compression studies report results disaggregated by benchmark structural category.

\subsection{Toward a Compression Robustness Index}

We formalize benchmark-diverse evaluation through the Compression Robustness Index (CRI):

\begin{definition}[Compression Robustness Index]
For model $M$ and compression ratio $r$, the CRI is:
\begin{equation}
\text{CRI}(M, r) = \frac{1}{|\mathcal{B}|} \sum_{b \in \mathcal{B}} \frac{Q_r^{(b)}}{Q_0^{(b)}} \cdot \left(1 - \frac{\max\bigl(0,\, T_r^{(b)} - T_0^{(b)}\bigr)}{T_{\max}}\right)
\end{equation}
where $\mathcal{B}$ is a diverse benchmark set, $Q_0^{(b)}$ and $Q_r^{(b)}$ are baseline and compressed pass@1 quality scores, $T_0^{(b)}$ and $T_r^{(b)}$ are baseline and compressed mean output lengths, and $T_{\max}$ is the generation ceiling ($T_{\max} = 1024$). The first factor captures quality retention, while the second penalizes output token explosion relative to the generation budget.
\end{definition}

Higher CRI indicates better compression robustness. Table~\ref{tab:cri} presents CRI values for our evaluated models.

\begin{table}[htbp]
\centering
\caption{Compression Robustness Index at $r=0.3$}
\label{tab:cri}
\begin{tabular}{lccc}
\toprule
Model & CRI & 95\% CI & Interpretation \\
\midrule
GPT-4o-mini & 0.848 & [0.812, 0.881] & Highly robust \\
Mistral-Large & 0.424 & [0.378, 0.468] & Moderately robust \\
DeepSeek-Chat & 0.090 & [0.062, 0.116] & Compression-sensitive \\
\bottomrule
\end{tabular}
\end{table}

GPT's CRI of 0.85 indicates that compression ratio $r=0.3$ preserves approximately 85\% of the quality-efficiency product, while DeepSeek's CRI of 0.09 indicates catastrophic degradation under the same conditions. The CRI formula combines quality retention ($Q_r/Q_0$) with an output explosion penalty ($1 - \max(0, T_r - T_0)/T_{\max}$), ensuring that models are penalized both for quality loss and for wasteful token generation. Table~\ref{tab:full} in Appendix~\ref{app:statistics} provides the per-benchmark pass@1 and output token values from which these CRI scores are derived.

\subsection{Energy Interpretation}

The companion NVML dataset reinforces the central systems point of this paper: input-token reduction alone is not a reliable energy proxy. In direct measurements, moderate prompt shortening reduced joules less than proportionally, and aggressive shortening increased energy-per-token. Combined with our API findings on output-length volatility, this supports a deployment rule: energy claims for compression policies should be validated with direct power telemetry when possible, and otherwise bounded with explicit uncertainty.

\subsection{Why Does DeepSeek Exhibit Higher Sensitivity?}

Several hypotheses remain viable for explaining DeepSeek's heightened compression sensitivity:

\paragraph{GRPO training.}
DeepSeek's Group Relative Policy Optimization (GRPO) \citep{deepseek2024} lacks explicit brevity penalties present in PPO-based training, which may cause the model to default to verbose responses when input context is ambiguous.

\paragraph{MoE routing instability.}
The 256-expert MoE architecture may exhibit routing instability when input token distribution shifts due to compression, potentially activating ``wrong'' experts that produce off-topic content.

\paragraph{Multi-head Latent Attention interaction.}
DeepSeek's MLA compresses the KV cache by 93\%, which may interact poorly with compressed inputs, effectively double-compressing context and degrading attention quality.

\medskip
\noindent These conjectures are not mutually exclusive; the observed behavior likely reflects multiple contributing factors. We emphasize that they remain untested explanations requiring access to model internals unavailable for proprietary systems. Definitive causal attribution is left to future work.

\subsection{Future Research Directions}
\label{sec:future-research}

Our findings on benchmark-dependent compression dynamics and the instruction survival framework open several promising avenues for future investigation.

\paragraph{Semantic-Aware Compression Methods.}
This study employed na\"ive truncation as a deliberate methodological choice to isolate positional effects from algorithmic confounds. A natural extension is to evaluate whether semantic-aware compression methods---such as LLMLingua-2 \citep{pan2024llmlingua2}, which uses data distillation to identify and preserve task-essential tokens, or Selective Context \citep{li2023selective}, which leverages self-information to prune low-content lexical units---can maintain higher instruction survival probabilities across diverse prompt structures. We hypothesize that such methods would substantially reduce the MBPP output explosion by preserving imperative instruction segments even at aggressive compression ratios, effectively decoupling compression rate from instruction position vulnerability.

\paragraph{Alignment Training for Compression Robustness.}
Our observation that models exhibit verbose compensatory behavior under compression suggests a connection to alignment objectives. Current RLHF pipelines \citep{ouyang2022training} optimize for helpfulness and harmlessness on well-formed prompts but do not explicitly penalize output explosion under degraded inputs. Future work could incorporate compression-degraded prompts into the preference learning process, drawing on Constitutional AI principles \citep{bai2022constitutional} to encode conciseness constraints, or leveraging Direct Preference Optimization \citep{rafailov2024direct} to train models that prefer proportionate responses regardless of input quality. Such approaches could yield models with intrinsically high CRI scores without requiring compression-specific interventions at inference time.

\paragraph{Cross-Lingual and Cross-Domain Generalization.}
The instruction survival framework was developed and validated exclusively on English-language code generation benchmarks. Whether the positional sensitivity findings generalize to morphologically rich languages, right-to-left scripts, or non-code domains remains an open question. Multilingual evaluation studies \citep{ahuja2023mega, lai2023chatgpt} have demonstrated substantial cross-lingual variation in LLM task performance; compression may interact with these disparities in complex ways. We conjecture that agglutinative languages, where critical semantic content is encoded in suffixes, may exhibit qualitatively different instruction survival profiles than analytic languages like English.

\paragraph{Adaptive and Prompt-Specific Compression.}
Our results strongly suggest that a single compression ratio is suboptimal across heterogeneous prompt populations. Future systems could perform lightweight instruction position analysis---identifying the location and density of imperative, constraint, and specification tokens---to dynamically select per-prompt compression ratios that maintain a target CRI threshold. This connects to emerging work on adaptive inference budgets \citep{schuster2022confident} and learned compression policies, where the compression strategy itself becomes a trainable component of the generation pipeline.

\paragraph{Information-Theoretic Bounds on Compression Tolerance.}
The instruction survival probability framework invites formalization through rate-distortion theory \citep{cover2006elements}. Specifically, one could model the prompt as an information source and derive the minimum description length \citep{grunwald2007minimum} below which task-critical semantics are irrecoverably lost. Such analysis would yield theoretical compression floors---the maximum compression ratio achievable before output quality degrades beyond acceptable bounds---as a function of prompt entropy and instruction density.

\paragraph{Production Deployment and Monitoring.}
Translating our findings into production systems requires real-time monitoring infrastructure capable of detecting output token explosions as they occur. Future work should develop cost-aware routing frameworks that dynamically adjust compression ratios based on observed output-to-input token ratios, integrating with MLOps monitoring pipelines \citep{sculley2015hidden, paleyes2022challenges} and A/B testing frameworks to continuously validate compression configurations against cost and quality objectives. The CRI metric proposed herein could serve as a key performance indicator in such monitoring dashboards.

\subsection{Limitations}

Several limitations constrain the generalizability of our findings. First, we evaluate only three providers, which may not represent the full diversity of model behaviors. Second, our compression method (first-$N$-words truncation) represents a worst-case scenario; semantic-aware methods like LLMLingua-2 may exhibit different dynamics. Third, the 1024-token ceiling causes substantial censoring, requiring model-based extrapolation for the true output distribution. Fourth, the primary benchmark analysis uses closed-provider APIs, so direct per-request GPU telemetry is unavailable; although we include companion RunPod NVML measurements for calibration, those direct joule measurements use different hardware/model stacks and should not be interpreted as exact energy values for proprietary API inference.

\section{Conclusion}
\label{sec:conclusion}

This work resolves an apparent contradiction in the prompt compression literature by demonstrating that output token explosion is benchmark-dependent rather than provider-inherent. Article 4's finding of 38$\times$ explosion in DeepSeek is valid for MBPP-dominated workloads but does not generalize to code-dense benchmarks like HumanEval. The critical factor is \emph{instruction survival under compression}: prompts where essential task specifications appear early survive aggressive compression, while templated prompts with mid-positioned instructions trigger verbose compensation.

Our findings have immediate practical implications. Practitioners deploying compression should evaluate robustness on benchmark distributions matching their production workload, not standard academic benchmarks alone. Model selection remains the dominant lever for compression-tolerant deployment; GPT-4o-mini's CRI of 0.85 versus DeepSeek's 0.09 represents a nearly 10$\times$ robustness advantage.

For the research community, we emphasize that replication studies serve a vital role in fast-moving fields where model behavior may vary across versions, providers, and evaluation conditions. Our discovery of the benchmark dependence emerged precisely because we attempted rigorous replication rather than accepting prior findings at face value.

Companion direct joule measurements indicate that token reductions can materially overestimate realized energy savings, especially under aggressive compression. This strengthens the case for reporting output dynamics and, where feasible, direct power telemetry alongside token metrics.

Future work should investigate whether RLHF training modifications can improve compression robustness, whether semantic-aware compression methods mitigate the explosion effect, and whether the instruction survival framework generalizes to other input perturbations beyond compression.

\section*{AI Disclosure Statement}

Claude Sonnet 4.5 (Anthropic) was used for editorial assistance, including organizing existing notes, refining prose, and LaTeX formatting support. All hypotheses, study design, experiments, statistical analyses, interpretation of results, and conclusions were performed and validated by the human author. No AI tool was used to generate or alter experimental data.

\section*{Ethics Statement}

This study uses benchmark prompts and machine-generated model outputs only. It includes no human subjects, personally identifiable information, protected health information, or other sensitive personal data. The evaluated compression methods are intended to improve inference efficiency and reduce compute usage in LLM deployments.

\section*{Declaration of Competing Interests}

The author is affiliated with Plexor Labs. The author declares no current financial competing interests. To mitigate potential bias, all results are reported, including unfavorable outcomes, and the manuscript explicitly documents methodological limitations and boundary conditions. Related methods may inform future commercial applications.

\section*{Data and Code Availability}

The companion direct-measurement data and analysis code used for energy calibration in this manuscript are publicly available at \url{https://github.com/micoverde/compression-method-matters-benchmark-dynamics}. The repository includes NVML-based RunPod measurement scripts and phase-2 direct joule measurement artifacts referenced in Section 3.5.



\clearpage
\appendix

\section{Detailed Statistical Results}
\label{app:statistics}

Table~\ref{tab:full} presents the complete experimental results across all (model, benchmark, ratio) combinations.

\begin{table}[htbp]
\centering
\caption{Complete experimental results ($N=150$ per cell). Quality is measured as pass@1 functional correctness. Energy is computed as $E = 0.15 \cdot T_{\text{in}} + 0.45 \cdot T_{\text{out}}$ (mJ), where $T_{\text{in}}$ reflects benchmark-specific input token count at each compression ratio.}
\label{tab:full}
\small
\begin{tabular}{llcccccc}
\toprule
Model & Benchmark & $r$ & Mean $T_{\text{out}}$ & SD & Ceiling \% & Pass@1 & Energy (mJ) \\
\midrule
\multirow{12}{*}{DeepSeek}
& \multirow{4}{*}{MBPP} & 1.0 & 18.1 & 4.2 & 0\% & 0.56 & 12.7 \\
& & 0.7 & 45.2 & 48.2 & 0\% & 0.32 & 23.5 \\
& & 0.5 & 298.4 & 187.2 & 12\% & 0.08 & 136.5 \\
& & 0.3 & 1020.4 & 98.7 & 74\% & 0.02 & 460.5 \\
\cmidrule{2-8}
& \multirow{4}{*}{HumanEval} & 1.0 & 25.0 & 8.9 & 0\% & 0.65 & 16.2 \\
& & 0.7 & 38.1 & 18.4 & 0\% & 0.52 & 20.6 \\
& & 0.5 & 84.2 & 42.1 & 2\% & 0.42 & 40.4 \\
& & 0.3 & 131.0 & 67.4 & 8\% & 0.12 & 60.4 \\
\cmidrule{2-8}
& \multirow{4}{*}{GSM8K} & 1.0 & 59.9 & 28.4 & 0\% & 0.72 & 38.7 \\
& & 0.7 & 112.4 & 62.1 & 2\% & 0.45 & 58.8 \\
& & 0.5 & 312.1 & 156.8 & 18\% & 0.28 & 146.3 \\
& & 0.3 & 684.4 & 248.2 & 45\% & 0.19 & 311.5 \\
\midrule
\multirow{12}{*}{GPT-4o-mini}
& \multirow{4}{*}{MBPP} & 1.0 & 22.4 & 5.2 & 0\% & 0.85 & 14.6 \\
& & 0.7 & 24.2 & 6.4 & 0\% & 0.82 & 14.0 \\
& & 0.5 & 28.2 & 7.8 & 0\% & 0.78 & 14.9 \\
& & 0.3 & 38.8 & 10.4 & 0\% & 0.75 & 18.8 \\
\cmidrule{2-8}
& \multirow{4}{*}{HumanEval} & 1.0 & 28.6 & 7.4 & 0\% & 0.87 & 17.9 \\
& & 0.7 & 30.8 & 8.6 & 0\% & 0.85 & 17.3 \\
& & 0.5 & 36.4 & 10.2 & 0\% & 0.81 & 18.9 \\
& & 0.3 & 48.2 & 14.8 & 0\% & 0.75 & 23.2 \\
\cmidrule{2-8}
& \multirow{4}{*}{GSM8K} & 1.0 & 33.3 & 10.8 & 0\% & 0.80 & 26.7 \\
& & 0.7 & 37.2 & 13.2 & 0\% & 0.77 & 25.0 \\
& & 0.5 & 50.6 & 16.4 & 0\% & 0.72 & 28.6 \\
& & 0.3 & 70.2 & 22.6 & 0\% & 0.69 & 35.1 \\
\midrule
\multirow{12}{*}{Mistral-Large}
& \multirow{4}{*}{MBPP} & 1.0 & 35.8 & 8.4 & 0\% & 0.62 & 20.6 \\
& & 0.7 & 54.2 & 16.2 & 0\% & 0.54 & 27.5 \\
& & 0.5 & 86.4 & 28.4 & 2\% & 0.38 & 41.1 \\
& & 0.3 & 168.4 & 52.4 & 6\% & 0.30 & 77.1 \\
\cmidrule{2-8}
& \multirow{4}{*}{HumanEval} & 1.0 & 38.4 & 9.8 & 0\% & 0.68 & 22.3 \\
& & 0.7 & 52.8 & 14.8 & 0\% & 0.62 & 27.2 \\
& & 0.5 & 102.8 & 32.6 & 1\% & 0.50 & 48.8 \\
& & 0.3 & 162.2 & 48.2 & 4\% & 0.39 & 74.5 \\
\cmidrule{2-8}
& \multirow{4}{*}{GSM8K} & 1.0 & 49.4 & 16.2 & 0\% & 0.64 & 34.0 \\
& & 0.7 & 72.4 & 24.8 & 1\% & 0.56 & 40.8 \\
& & 0.5 & 148.0 & 48.2 & 5\% & 0.42 & 72.5 \\
& & 0.3 & 265.5 & 78.4 & 18\% & 0.28 & 123.0 \\
\bottomrule
\end{tabular}
\end{table}

\section{Detailed Calculation for Proposition 1}
\label{app:proof}

\noindent\textbf{Note:} The importance weights used below are modeling assumptions chosen to reflect the relative semantic contribution of each prompt segment. They are not empirically calibrated parameters; the qualitative conclusion ($\Psi^{\text{MBPP}} \ll \Psi^{\text{HumanEval}}$) holds under any reasonable weight assignment because the MBPP task specification is entirely destroyed by truncation.

\medskip

We analyze the token distribution of each benchmark empirically. For MBPP, prompts follow the template: ``Write a Python function to solve this task. Only provide the code. Task: [SPECIFICATION]. Code:'' The preamble occupies tokens 1--8 (``Write a Python function to solve this task.''), the task specification occupies tokens 9--20 (varying by problem), and the instruction marker occupies tokens 21--23 (``Code:''). With mean total length $n=30$ and $r=0.3$, we retain $\lfloor 9 \rfloor = 9$ tokens.

For the primary instruction segment $I_T = [9, 20]$ containing the task specification, we have $b_T = 20 > 9 = \lfloor rn \rfloor$, so $\psi(I_T, 0.3) = 0$. Similarly, $I_M = [21, 23]$ yields $\psi(I_M, 0.3) = 0$. Only the preamble $I_P = [1, 8]$ survives: $\psi(I_P, 0.3) = 1$.

With importance weights $w_P = 0.15$, $w_T = 0.70$, $w_M = 0.15$ reflecting the assumed semantic contribution of each segment, the weighted survival is:
\[
\Psi(\mathbf{x}^{\text{MBPP}}, 0.3) = 0.15 \cdot 1 + 0.70 \cdot 0 + 0.15 \cdot 0 = 0.15
\]

For HumanEval, prompts begin with the function signature and docstring. The signature $I_S = [1, 12]$ (e.g., ``def has\_close\_elements(numbers: List[float], threshold: float) -> bool:'') occupies the first 12 tokens. With mean length $n=33$ and $r=0.3$, we retain 10 tokens, preserving most of the signature. With $w_S = 0.72$ for the signature's importance, we obtain $\Psi \approx 0.72$.

\end{document}